\documentclass[letterpaper, 10 pt, conference]{ieeeconf}  

\IEEEoverridecommandlockouts                              

\usepackage[numbers]{natbib}
\usepackage{graphicx}
\usepackage{amsmath}
\usepackage{hyperref}
\usepackage{algpseudocode}
\usepackage{algorithm}
\usepackage[algo2e]{algorithm2e}
\usepackage{multirow}
\usepackage[normalem]{ulem}
\usepackage{booktabs}
\usepackage{gensymb}
\usepackage{lipsum}

\usepackage{amsmath,amsfonts,bm}
\usepackage{caption}
\usepackage{subcaption}

\def\eqref#1{equation~\ref{#1}}

\def\1{\bm{1}}

\DeclareMathAlphabet{\mathsfit}{\encodingdefault}{\sfdefault}{m}{sl}
\SetMathAlphabet{\mathsfit}{bold}{\encodingdefault}{\sfdefault}{bx}{n}

\newcommand{\xxnote}[3]{}
\ifx\hidenotes\undefined
  \renewcommand{\xxnote}[3]{\color{#2}{#1: #3}}
\fi

\hypersetup{
    colorlinks=true,
    linkcolor=blue,
    filecolor=magenta,      
    citecolor=blue,
    urlcolor=blue,
}

\usepackage[font={small}]{caption}
\renewcommand\thefigure{\arabic{figure}}
\newcommand{\method}{\textsc{TAVI}}
\newcommand{\website}{\url{https://see-to-touch.github.io/}}

\title{\LARGE \bf
See to Touch: Learning Tactile Dexterity through Visual Incentives
}

\author{
Irmak Guzey$^{1,\dagger}$ \qquad Yinlong Dai$^1$ \qquad Ben Evans$^1$ \qquad Soumith Chintala$^2$  \qquad Lerrel Pinto$^1$
\\ \\ New York University$^1$, Meta AI Research$^2$
\\ \\ { \tt \href{https://see-to-touch.github.io/}{see-to-touch.github.io}}
\thanks{$^{\dagger}$Correspondence to \texttt{irmakguzey@nyu.edu}.}
}

\newboolean{include-notes}
\setboolean{include-notes}{true}

\begin{document}


\maketitle
\thispagestyle{empty}
\pagestyle{empty}

\begin{abstract}
Equipping multi-fingered robots with tactile sensing is crucial for achieving the precise, contact-rich, and dexterous manipulation that humans excel at. However, relying solely on tactile sensing fails to provide adequate cues for reasoning about objects' spatial configurations, limiting the ability to correct errors and adapt to changing situations. In this paper, we present Tactile Adaptation from Visual Incentives (TAVI), a new framework that enhances tactile-based dexterity by optimizing dexterous policies using vision-based rewards. First, we use a contrastive-based objective to learn visual representations. Next, we construct a reward function using these visual representations through optimal-transport based matching on one human demonstration. Finally, we use online reinforcement learning on our robot to optimize tactile-based policies that maximize the visual reward. On six challenging tasks, such as peg pick-and-place, unstacking bowls, and flipping slender objects, TAVI achieves a success rate of 73\% using our four-fingered Allegro robot hand. The increase in performance is 108\% higher than policies using tactile and vision-based rewards and 135\% higher than policies without tactile observational input. Robot videos are best viewed on our project website: \website{}.

\end{abstract}


\section{Introduction}
\label{sec:intro}

Dexterity has played a crucial role in human development, enabling us to create and utilize tools effectively~\cite{KARAKOSTIS20211317}. Although two-fingered grippers have been extensively studied in the field of robotics~\cite{ferrari1992planning, 10.1007/978-3-030-33950-0_33, pinto2016supersizing}, they inherently lack the physical capabilities required for performing dexterous tasks that necessitate fine-grained manipulation at the fingertips. These additional capabilities facilitate a wider range of tasks in real-world scenarios; however, they also result in a higher dimensional actions. Furthermore, due to visual occlusion during such manipulation processes, effective utilization of tactile data becomes vital -- an aspect that remains understudied in the context of dexterity.

\begin{figure}[t]
    \centering
    \includegraphics[width=\linewidth]{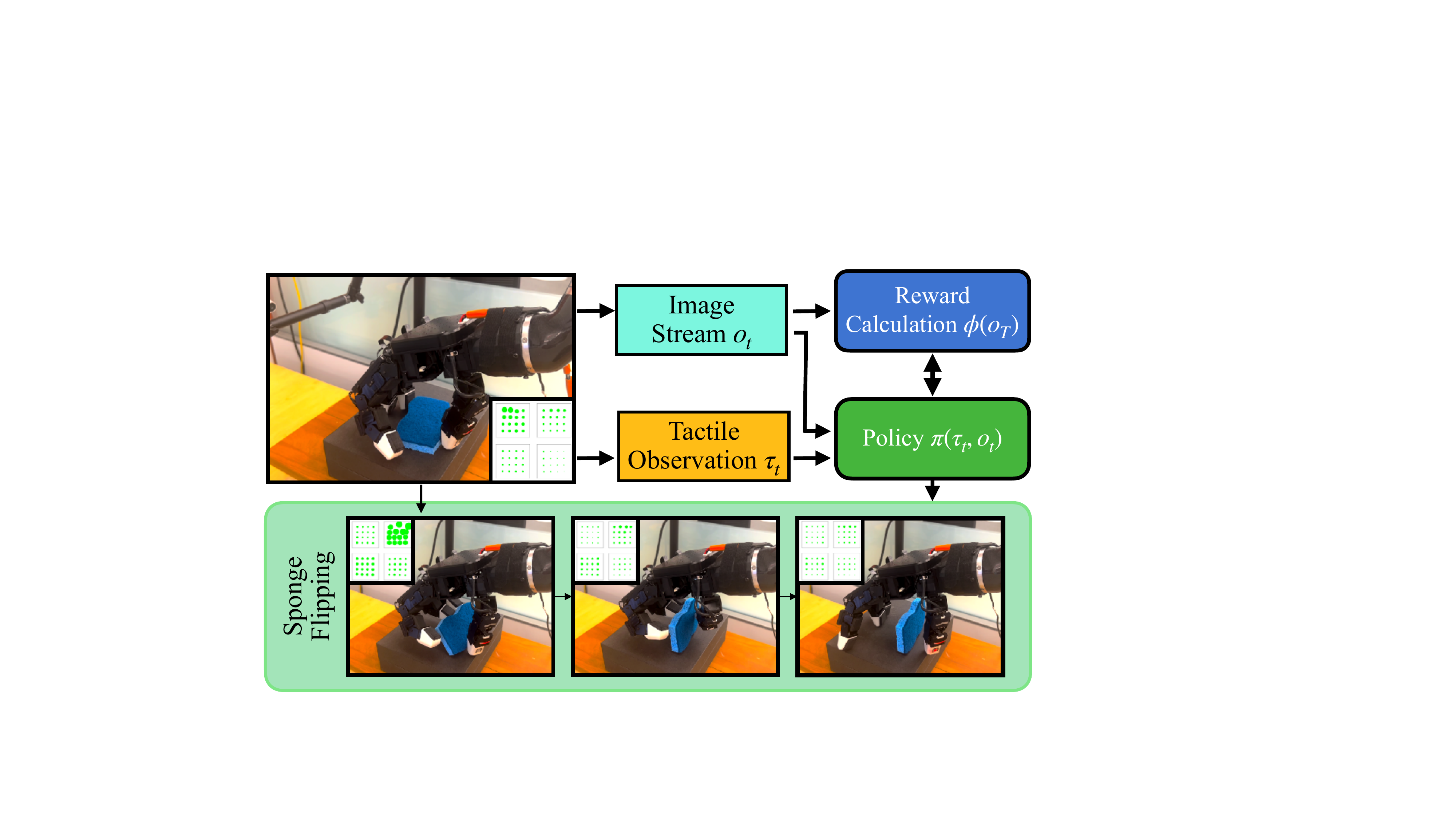}
    \caption{\method{} learns dexterous policies through online learning. Both tactile and image is used to retrieve action while only image is used for reward calculation.}
    \label{fig:figure_1}
\end{figure}

To train dexterous policies, several frameworks have been proposed, ranging from model-based control, in which models of the robot and object are used to optimize control behavior~\cite{Rajeswaran2018, Nagabandi2019}, to simulation-to-reality transfer (sim2real), where a policy is trained in a simulator and then transferred to the real world~\cite{openai2019learning, nvidia2022dextreme}. While the latter methods demonstrate impressive results, they require the ability to simulate sensory observations during manipulation. This becomes problematic when using rich tactile sensing, as modeling uncalibrated skin sensing is an open research problem in itself~\cite{lee2020calibrating}. Consequently, much of the prior work in multi-fingered dexterity relies either exclusively on visual feedback or on weaker binary-touch signals~\cite{yin2023rotating}.

To address the challenges associated with modeling dexterous behavior, recent approaches have focused on imitation-based methods. In these approaches, humans first teleoperate robots to collect demonstrations of dexterous behavior~\cite{arunachalam2022dexterous, arunachalam2022holodex, guzey2023dexterity}. Then, offline imitation learning is used to obtain policies that fit these demonstrations. Dexterous policies trained using this approach can solve a wide range of tasks, from reorienting objects to precise picking. Importantly, since policies are trained on real observational data, they readily scale to skin-based tactile data, which is otherwise difficult to model. However, offline imitation is not a silver bullet. First, it requires the collection of significant amounts of difficult-to-collect demonstration data. Second, it needs the demonstrations to densely cover object configurations used in evaluation. Third, it does not have any mechanism to recover from errors during execution.

In this work, we present Tactile Adaptation from Visual Incentives (\method{}), a new framework for tactile-based dexterity that requires only one successful demonstration, can generalize to new object configurations, and can learn to correct behaviors from failures. The key insight in \method{} is to continuously adapt the dexterous policy by improving the Optimal-Transport (OT) match between sensory observations generated by the policy and those generated through human demonstrations. The adaptation algorithm is built on prior work in inverse reinforcement learning (IRL)~\cite{haldar2023teach}, where the matching score corresponds to rewards and policy optimization is done through RL.

However, unlike prior work in IRL, where observations can directly provide crisp signals of task completion, the use of tactile observations poses a unique challenge. Tactile signals often lack the necessary cues to reason about the spatial location of objects. For instance, the tactile signal obtained from picking up a slender object is very similar to that of pinch grasping. To address this, we only use visual cues to determine reward, which in turn provides a strong incentive for tactile-based policy learning. Moreover, to improve the quality of visual rewards, we use a novel contrastive learning objective that augments a time-contrastive loss with proprioceptive movements.

We experimentally evaluate \method{} on six contact-rich, dexterous tasks such as opening a mint box, unstacking bowls, and flipping slender objects. Through an extensive study, we present the following key insights:
\begin{enumerate}
    \item \method{} improves upon prior state-of-the-art work in dexterous imitation~\cite{guzey2023dexterity} with an average of $5.5\times$ improvement in success rate given 30 minutes of online interactions. This represents the first framework to learn dexterous policies from online tactile-based interactions (Section~\ref{sec:experiments:main}).
    \item Visual representations learned through our contrastive learning scheme achieve approximately 56\% improvement in four of our tasks over prior representation methods on dexterous manipulation trajectories (Section~\ref{sec:experiments:enc}).
    \item Ablations on different representation modules and sensor combinations show that the design decisions in \method{} are crucial for high performance (Sections~\ref{sec:experiments:main}, \ref{sec:experiments:enc}). 
\end{enumerate}

Robot videos generated by \method{} are best viewed on our website: \website{}.

\section{Related Works}
\label{sec:related}


\paragraph{Dexterous manipulation and tactile sensing} Control of dexterous, multi-fingered robots has been of longstanding interest to the field~\cite{Ciocarlie2007DexterousGV, kumar2014real, Shigemi2018}. A recent approach is to learn a policy in simulation and transfer to the real world, which requires extensive randomization and does not simulate fine-grained touch sensors~\cite{openai2019learning, nvidia2022dextreme, yin2023rotating, qi2023general}.
Earlier work focuses on physics-based models of grasping \cite{okamura2000overview, odhner2014compliant} to compute grasp stability from motor torque readings. Unfortunately, these methods are susceptible to noise due to the inherent interconnection between motor sensors and controllers.
To mitigate this coupling, a number of tactile sensors have been created to endow robots with touch~\cite{https://doi.org/10.48550/arxiv.2106.08851, bhirangi2021reskin, alspach2019soft}. One such sensor, GelSight, has been used extensively for tasks like object classifcation~\cite{patel2021digger}, measuring surface properties~\cite{dong2017improved}, in-hand rotation~\cite{qi2023general} and pose estimation~\cite{kelestemur2022tactile}. Due to GelSight's difficulty to cover an entire multifingered hand, `skin'-like sensors~\cite{8858052} have been developed. These sensors can cover the entire hand, giving high-resolution tactile information that can aid learning dexterous policies.

\paragraph{Learning from tactile data} 
To leverage high-resolution readings from tactile sensors, a number of learning-based approaches have been used to solve tasks with two-fingered robot grippers ~\cite{10.1007/978-3-030-33950-0_33, calandra2018more, zambelli2021learning,https://doi.org/10.48550/arxiv.1910.02860, wang20183d}. These methods require a large amount of task-oriented data and are not applied to multi-fingered hands. Most similar to our work, \textsc{T-Dex}~\cite{guzey2023dexterity} learns a tactile representation for an entire hand by using self-supervision on a large, task-agnostic play dataset. \method{} uses the pretrained tactile encoder from \textsc{T-Dex} for our tasks.

\paragraph{Representation learning for visual observations}

Learning meaningful, low-dimensional representations with limited or no data labels is an active area of interest in computer vision~\cite{simclr, moco2, grill2020bootstrap, swav}. These techniques aim to optimize an auxiliary objective that results in representations that are good for downstream tasks. Some tasks include maintaining consistency between augmentations of the same image~\cite{grill2020bootstrap}, reconstruction of patches~\cite{he2021masked}, and making sure similar examples are close to one another~\cite{oord2019representation}. This has been successfully applied on computer vision benchmarks due to the availability of large amounts of unlabeled data~\cite{bardes2021vicreg, dwibedi2021little, assran2023selfsupervised}.
Because of the limited availability of labeled data in robotics, unsupervised and semi-supervised representation learning techniques have grown in popularity for tasks like manipulation~\cite{manuelli2020keypoints} and visual imitation~\cite{young2020visual, pari2021surprising}. For our experiments, we use an InfoNCE-style~\cite{oord2019representation} loss using time-contrastive~\cite{Sermanet2017TCN} pairs to learn visual representations.

\paragraph{Online adaptation and imitation learning}


Imitation learning (IL) is has been effective for solving real-world tasks~\cite{argall2009survey, hussein2017imitation}. The simplest form, Behavior Cloning (BC) learns policies from offline expert demonstrations and it has been effective especially with large datasets~\cite{pomerleau1998autonomous, torabi2019recent}. However, BC struggles with out-of-domain scenarios~\cite{ross2011reduction}. Inverse reinforcement learning (IRL) estimates expert reward functions, enhancing policy performance but at the cost of sample efficiency~\cite{kostrikov2018discriminator}. Many works have sought to improve the efficiency of IRL~\cite{kostrikov2018discriminator, fu2017learning,xiao2019wasserstein} and to extend it to the visual imitation setting ~\cite{haldar2022watch, cetin2021domain,toyer2020magical,rafailov2021visual}. Our work leverages optimal transport IRL for efficient policy learning from visual inputs~\cite{haldar2023teach}.

\section{Background}

\method{} builds on several technical ideas in constrastive learning and optimal-transport imitation:

\subsection{Constrastive Self-Supervised Learning}
\label{sec:infonce}
Self-supervised learning (SSL) seeks to learn compact representations for high-dimensional observations, such as images, to be used in downstream tasks. Contrastive methods for SSL seek to move representations between ``positive'' samples close together while moving ``negative'' samples further from one another. 

InfoNCE~\cite{oord2019representation} is a commonly used loss function employed in contrastive learning that distinguishes positive and negative pairs based on their density ratio. For an observation and its positive pair $o_t, o_t^+$ and set of $n$ negative observations $\mathcal{D}=\{o_1, \ldots, o_n\}$, resulting in latents $z_t, z_t^+$ and $\{z_1, \ldots z_n\}$, the loss is defined as:
 \begin{equation}
\mathcal{L}_\text{NCE}(z_t, z_t^+, \{z_1, \ldots z_n\}) =
-\mathbb{E}_{\mathcal{D}}\left[\log{\frac{h(z_{t}, z_t^+)}{\sum_{i=1}^n h({z}_{t}, {z}_i)}}\right]
\label{eq:cpc}
\end{equation}
where $h(x,y)=\exp(x\cdot y)$.
Maximizing this loss causes the model to assign higher probabilities to positive pairs while pushing apart negative, resulting in discriminative representations. 

\subsection{Optimal-Transport Imitation Learning}
\label{sec:fish}
Imitation learning seeks to find a policy from expert demonstrations. Recent methods~\cite{haldar2022watch,cohen2022imitation} have used optimal-transport to efficiently imitate expert trajectories from images. One of these methods, FISH~\cite{haldar2023teach}, takes a weak base policy and an encoder that maps from high dimensional observations to a low dimensional latent space, and learns a residual policy that corrects the base policy by producing corrective offsets. It does this by using an optimal-transport-based reward function between an expert trajectory and the robots executed trajectory. Formally, given an expert trajectory $\mathcal{T}^e=\{o_1^e, \ldots, o_T^e\}$ and an observed robot trajectory $\mathcal{T}^r=\{o_1^r, \ldots, o_T^r\}$, latent representations for each $\{z_1^e, \ldots, z_T^e\}$, $\{z_1^r, \ldots, z_T^r\}$ are computed using the given encoder. A pairwise cost matrix between the two representations, $C$, can then be formed where $C_{ij}$ corresponds to the cost of moving $z_i^e$ to $z_j^r$. Optimal-transport finds the transport plan $\mu^*$ that best matches $\mathcal{T}^e$ and $\mathcal{T}^r$, where $\mu^*_{ij}$ is the score of the match between the $i$th representation from the expert and $j$th representation from the robot. The optimal-transport reward is computed as 
\begin{equation}
r^{\text{OT}}(o_t^r) = - \sum_{t'=1}^{T}C_{t,t^{'}}\mu^*_{t,t'}
\end{equation}
This allows for comparing behaviors in a time-invariant manner. If $\mathcal{T}^r$ exactly matched $\mathcal{T}^e$, the cost would be zero everywhere and the reward would be maximized. If our robot trajectory was offset, say by repeating the first observation $\mathcal{T}^r=\{o_1^e, o_1^e, \ldots o_{T-1}^e\}$, we would only be lightly penalized because the optimal-transport would find good matches between adjacent observations. DDPG~\cite{lillicrap2015continuous} is used to maximize this reward function, resulting in similar behavior to the expert.

\section{Tactile Adaptation through Visual Incentives (\method{})}
\label{sec:method}

First, we collect data on a robot hand equipped with skin-based tactile sensors. Expert demonstrations are collected using the \textsc{Holo-Dex} framework (Section \ref{sec:tavi:setup}). Next, we must obtain visual representations for OT reward calculation. This is done in a self-supervised manner with a modified InfoNCE loss (Section \ref{sec:tavi:rep}). Finally, we train a policy online to imitate the expert demonstration using an OT-based reward function with features from the learned visual encoder (Section \ref{sec:tavi:policy}).

\subsection{Robot Setup and Expert Data Collection}
\label{sec:tavi:setup}
Our robot is an arm-hand system with a 6-DOF Jaco arm and a 16-DOF AllegroHand (see Figure~\ref{fig:figure_1}~(a)). The hand is fitted with 15 XELA uSkin tactile sensors~\cite{8307485}, each with a 4x4 tri-axial force reading, and we place an RGB camera in the scene to capture visual information.
We collect data using the \textsc{Holo-Dex} framework~\cite{arunachalam2022holodex}, which uses a VR headset to track hand pose and re-targets to a similar pose on the robot morphology. During data collection, we record the position and orientation of the arm's end effector, $s^\text{arm}$, the positions of all of the joints on the hand, $s^\text{hand}$, as well as the tactile and image information, $\tau, o$. Since the robots and sensors all return data at different frequencies, we align the data using the collected timestamps, and combine the robot states $s_t=s_t^\text{arm} \oplus s_t^\text{hand}$ to produce aligned tuples $(s_t, \tau_t, o_t)$. Similar to~\cite{guzey2023dexterity}, we subsample the data to 10Hz and remove transitions where the cumulative movement is below 1 cm.

\subsection{Representation Learning for Vision and Tactile Observations}
\label{sec:tavi:rep}

In order to mitigate the need for explicit state estimation, we use self-supervised learning to learn a mapping from high dimensional observations to a lower dimensional latent state (see Section~\ref{sec:infonce} for more details). The image encoder, which maps images $o_t$ to latents $z_t$, is trained on demonstration data for the task and uses a combination of two losses. The first is the InfoNCE~\cite{oord2019representation} loss trained using nearby observations as positive examples, following the methodology of Time-Contrastive Networks~\cite{Sermanet2017TCN}. The second loss predicts the change in robot state between nearby observations using a small mlp head, $\hat{\Delta}(z_t, z_{t+k})$. The change loss function is $\mathcal{L}_\Delta(s_t, s_{t+k}, z_t, z_{t+k}) = ||s_{t+k} - s_t - \hat{\Delta}(z_t, z_{t+k})||$ and differs from an inverse model~\cite{Agrawal2016} in that we predict a change in state over multiple steps instead of a single action. Our final loss function is
$$\mathcal{L} = \mathcal{L}_\text{NCE}(z_t, z_{t+k}, \{z_1, \ldots z_n\}) + \lambda \mathcal{L}_\Delta(s_t, s_{t+k}, z_t, z_{t+k})$$
For computational efficiency, we use the same observations for both the positive samples and to predict the change in joint angles, setting $k=5$ for all our experiments. We scale the losses to be approximately the same magnitude. For the tactile encoder, we download and use a pretrained tactile encoder that was trained on 2.5 hours of tactile-based play data using self-supervised learning~\cite{guzey2023dexterity}.

\begin{figure*}[t]
    \centering
    \includegraphics[width=\linewidth]{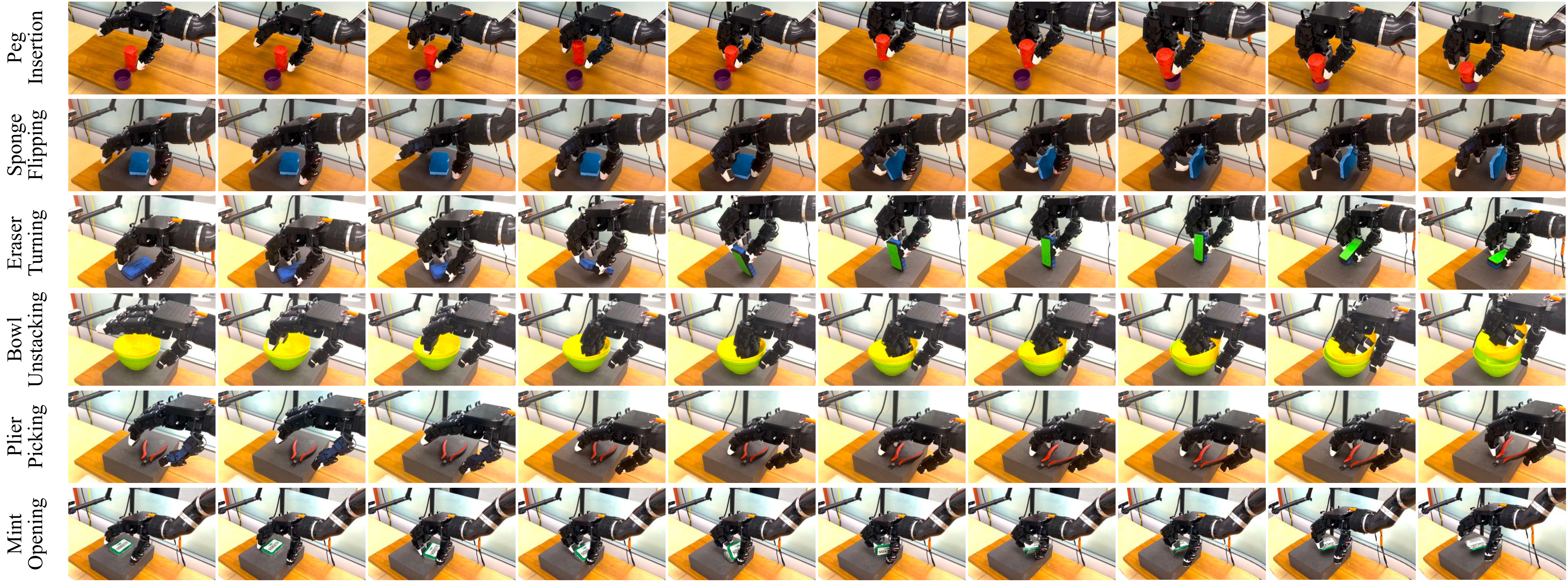}
    \caption{Rollouts of trained policies from \method{} on six tasks. Videos are best viewed on our website \website{}.}
    \label{fig:rollouts}
    \vspace{-0.1in}
\end{figure*}

\subsection{Policy Learning through Online Imitation}
\label{sec:tavi:policy}
We utilize the FISH~\cite{haldar2023teach} imitation algorithm on a single demonstrated trajectory to learn a policy (see Section~\ref{sec:fish} for more details). The base policy we choose is simply an open loop rollout of the expert demonstration, executing the previously executed actions in sequence. This policy completely fails if the environment is not the same as in the expert demonstration, but it serves as a decent base from which to learn a residual policy.
The offset policy receives both $o_t$ and $\tau_t$, which are augmented with random resized crops before passing to the visual and tactile encoders. Crucially, we only use visual information, $o_t$ to calculate the optimal-transport reward. We find that including the tactile information in the reward results in convergence to sub-optimal solutions like activating the touch sensors by pinching the fingers together. Details of the reinforcement learning (RL) agent that is used in \method{} can be found in Appendix \ref{sec:model_details}.

\paragraph{Final frame matching for rewards} We differ from \textsc{FISH} in that we do not use the entire length of trajectories to calculate the reward. Instead, we match the last 10 frames of the robot trajectory to the last frame of the expert trajectory. While this does not give us immediate feedback if the executed trajectory differs from the expert, the sparse reward of distance to the expert's final frame has enough signal to learn to complete the tasks. Including all of the frames results in matching starting robot frames to ending expert frames, and vice-versa, preventing the policy from completing the task. We further explain this behavior in Section \ref{sec:experiments:frames} and show an illustrative figure in Appendix \ref{sec:reward_details}. 

\paragraph{Exploration strategy} Since the method learns a residual policy, we can naturally enable or disable learning on subsets of the action space, i.e., we can explore along only the dimensions of the action space that we need to. We detail which parts of the action space are enabled in Section \ref{sec:experiments}.
To effectively explore the space, we use additive OU noise~\cite{PhysRev.36.823}, which prevents motor jitter.

\section{Experimental Evaluation}
\label{sec:experiments}


We experimentally evaluate \method{} to answer the following questions:
(a) How well does \method{} perform on dexterous tasks? (b) Does the contrastive encoder improve visual representation quality? (c) How well does \method{} generalize to unseen objects?

\subsection{Task Descriptions}
We experiment with six dexterous tasks that require precise control, visualized in Figure \ref{fig:rollouts}.

\paragraph{\textbf{Peg Insertion}}  The robot must locate and pick up a peg before inserting it into a cup on the table. The cup stays in the same position for all trials. We learn a residual policy on the base joints of all of the fingers.
\paragraph{\textbf{Sponge Flipping}} The robot must find a sponge lying flat on the table and manipulate it to balance it on its side. This is challenging since minor errors will result in the object tipping over. We learn a residual policy on the base of the thumb, index, and middle fingers.
\paragraph{\textbf{Eraser Turning}} The robot must pick a whiteboard eraser lying flat and rotate it 180$^{\circ}$ to lie flat on its opposite face. We learn a residual policy on the last two joints of the middle finger and the top three tip joints of the thumb.
\paragraph{\textbf{Bowl Unstacking}} The robot must locate a stack of bowls and remove a bowl from the stack. This task requires shear force to separate the bowls from one another. The residual policy learns the side-to-side offset of the arm end effector position and the thumb base joint.
\paragraph{\textbf{Plier Picking}}  The robot must locate and pick up a pair of pliers on the table. This task is especially difficult due to the precision required when placing the fingers. The residual policy learns offsets for the last pointer joint, the base and tip joints of the middle finger, and the two base joints of the thumb.
\paragraph{\textbf{Mint Opening}}  The robot must locate and open a metal mint box by using tips of the thumb, index and middle finger. This task requires robot to stabilize the box with the middle and thumb fingers and carefully opening the top of the mint box.

\begin{table*}[t!]
\caption{Success rates of \method{} and our baselines for evaluations run on the Allegro hand.}
\label{tab:success}
\centering
\begin{tabular}{lcccccc}
                & \multicolumn{1}{l}{\textsc{T-Dex~\cite{guzey2023dexterity}}} & \multicolumn{1}{l}{BC-BeT~\cite{shafiullah2022behavior}} & \multicolumn{1}{l}{Tactile Only} & \multicolumn{1}{l}{Image + Tactile Reward} & \multicolumn{1}{l}{AVI~\cite{haldar2023teach}} & \multicolumn{1}{l}{\method{}} \\ \hline
Peg Insertion   & 2/10                      & 0/10         & 6/10        & 6/10                                       & 6/10                    & \textbf{8/10}            \\
Sponge Flipping & 1/10                      & 0/10         & \textbf{8/10}          & 4/10                                       & 3/10                    & \textbf{8/10}            \\
Eraser Turning  & 2/10                      & 0/10         & 0/10          & 2/10                                       & 0/10                    & \textbf{5/10}            \\
Bowl Unstacking & 1/10                      & 0/10         & 5/10          & 0/10                                       & 3/10                    & \textbf{9/10}            \\
Plier Picking   & 0/10                      & 0/10         & 4/10          & 4/10                                       & 6/10                    & \textbf{7/10}            \\ 
Mint Opening    & 4/10                      & 0/10         & 0/10          & 5/10                                       & 1/10                    & \textbf{7/10}            \\
\hline
Avg. Success Rate    & 0.16                     & 0.0     &   0.38             & 0.35                                     & 0.31                    & \textbf{0.73}         
\end{tabular}
\end{table*}

\subsection{Baselines and Evaluation Metrics}
We study the effectiveness of our method and compare against the baselines described below:
\begin{enumerate}
    \item \textbf{\textsc{T-Dex}}~\cite{guzey2023dexterity}: We implement and run a state-of-the-art method for learning dexterous policies that utilizes self-supervised image and tactile encoders with nearest neighbors imitation. For the sake of fairness, we use the same image encoder used in \method{}.
    \item \textbf{BC-BeT}~\cite{shafiullah2022behavior}: We implement and run a state-of-the-art behavior cloning method Behavior Transformers. Again, we use the same encoders as our method and do not update the encoder parameters during training.
    \item \textbf{Tactile Only}: We only use the tactile representations for inputting to the policy and the calculating the optimal transport reward. This studies our choice of having image representations in \method{}.
    \item \textbf{Tactile and Image Reward}: To study the effect of our choice of reward function, we experiment with calculating the optimal transport reward from both tactile and visual features. We concatenate both features at each timestep and then run OT matching on it. 
    
    \item \textbf{No Tactile information (AVI)} ~\cite{haldar2023teach}: To study the value of tactile feedback, we train our method without tactile information given to the policy. While the reward calculation is the same as our main method, the policy must infer contact from vision alone.
\end{enumerate}

\paragraph{Evaluating robot performance} We allow all online imitation methods to train online with one expert demonstration until the reward converges or for up to 30 minutes. We evaluate all methods by running 10 rollouts with varying position and orientation of the manipulated objects. For fairness, we use the same 10 positions for each method. 
\paragraph{Evaluating visual representations} In order to evaluate our approach in learning visual representations we have run robot experiments on 4 of our tasks with 5 different set of encoders. In addition to the vision encoder in \method{}, we evaluate the following encoders on our framework:
\begin{enumerate}
    \item \textbf{Contrastive Only}~\cite{oord2018representation}: Similar training framework to \method{} but the loss doesn't include the change loss function. So the final loss only includes the InfoNCE loss between the temporal frames.  $$\mathcal{L} = \mathcal{L}_\text{NCE}(z_t, z_{t+k},\{z_1, \ldots z_n\})$$
    \item \textbf{Joint Difference}~\cite{brandfonbrener2023inverse}: Similar training framework to \method{} but the loss doesn't include the InfoCNE loss function. So the final loss only includes the change loss function. $$\mathcal{L} =  \lambda \mathcal{L}_\Delta(s_t, s_{t+k}, z_t, z_{t+k})$$
    \item \textbf{BYOL}~\cite{grill2020bootstrap}: We use the self-supervised learning algorithm; Bootstrap Your own Latent (BYOL) to train encoders on the task data.
    \item \textbf{BC}~\cite{pomerleau1989alvinn}: We receive visual representations from a simple 3-layered CNN and map them to actions applied during demonstrations. We train this end-to-end on the task data for each task. 
    \item \textbf{Pretrained}: We use a Resnet-18~\cite{he2016deep} with weights pre-trained on the ImageNet~\cite{deng2009imagenet} task with no finetuning.
\end{enumerate}


\subsection{How well does \method{} perform on dexterous tasks?}
\label{sec:experiments:main}

In Table \ref{tab:success} we report the success rates of \method{} and baselines. We see that BC-BeT is unable to complete any of the tasks, quickly going out of distribution and failing to recover. \textsc{T-Dex} is only able to solve at most 4 of the 10 runs, failing because it is unable to update the policy when the object has moved out of the demonstration set. While the combined image and tactile reward or tactile-only are able to solve more tasks than \textsc{T-Dex}, the noise introduced into the reward from the tactile information halves the success rate when compared to \method{}, highlighting the importance of computing rewards from visual information only. 

AVI almost matches the performance of our method on the peg insertion and plier picking tasks, but is unable to succeed at all on eraser turning and mint opening and has degraded performance on the other tasks. Neither peg insertion nor plier picking require precise force feedback to succeed, while eraser turning, mint opening, sponge flipping, and bowl unstacking all require a level of precision, taking care not to exert too much force on the manipulated object(s). These results underscore the importance of incorporating tactile feedback into dexterous policies.

We showcase the \method{} training rollouts and the corresponding OT rewards for each task in  Appendix \ref{sec:training_rollouts}.

\begin{table}[t!]
\caption{Success rates of different visual representations on \method{} learning framework}
\centering
\begin{tabular}{@{}ccccc|c@{}}
\toprule
Encoder               & Plier   & Bowl & Sponge  & Peg & Average                                                     \\ \midrule
TAVI                  & \textbf{7/10} & \textbf{9/10}  & 7/10 & \textbf{8/10} & \textbf{7.75/10}                     \\
Contrastive Only~\cite{oord2018representation}      & 0/10 & 7/10     & 2/10  & 7/10 & 4/10                                                        \\
Joint Difference~\cite{brandfonbrener2023inverse}      & 4/10 & 7/10     & 6/10 & 5/10 & 5.5/10                                                       \\
BYOL~\cite{grill2020bootstrap}                 & 6/10 & 5/10     & \textbf{9/10} & 6/10 & 6/10                                                \\
BC~\cite{pomerleau1989alvinn}                  & 0/10 & 6/10     & 3/10  & 6/10 & 4/10                                                        \\
Pretrained            & 5/10 & 6/10     & 2/10  & 5/10 & 4.5/10                                                      \\ \bottomrule
\end{tabular}
\label{tab:enc_results}
\vspace{-0.2in}
\end{table}

\subsection{Does the contrastive encoder improve visual representation quality?}
\label{sec:experiments:enc}

We report the success rates of experimented visual representations in Table \ref{tab:enc_results}. We observe that \textit{Pretrained}, \textit{Contrastive Only} and \textit{BC} encoders are not performing well due to failure in capturing the configuration between the object and the robot hand. We observe that \textit{Joint Difference Only} baseline performs relatively well since the encoder learns how to differentiate the impact of the object and actions to the hand pose but not as high as \method{} since it's lacking the temporal information coming from the contrastive loss. \textit{BYOL} training on the task data has been our most successful baseline after \method{}, we believe this is due to BYOL augmentations being able to force the visual representations to focus on the manipulation. Given the highest score in these experiments we choose to train our encoders with the combined contrastive and joint-prediction loss. 

\subsection{How does the number of frames included in the reward impact the results?}
\label{sec:experiments:frames}

As was mentioned in the Section \ref{sec:tavi:policy}, we only match the last frame of the expert demonstration and the last 10 frames of the robot trajectory in OT reward calculation.  We ran additional experiments with all of our tasks where we include all of the frames of both the robot and the expert trajectory and evaluated the policy with a similar evaluation setup. We show the results of this experiment in Table \ref{tab:frame_results}. 

\begin{table}[h]
\caption{Success rates of our learning framework with variant number of frames included in reward calculation.}
\begin{tabular}{ccccccc}
\toprule
Frames          & Bowl          & Peg           & Sponge        & Mint          & Plier         & Eraser        \\ \midrule
All frames inc. & 4/10          & 5/10          & 6/10          & 0/10          & 3/10          & 0/10          \\
TAVI            & \textbf{9/10} & \textbf{8/10} & \textbf{8/10} & \textbf{7/10} & \textbf{7/10} & \textbf{5/10} \\ \bottomrule
\end{tabular}
\label{tab:frame_results}
\end{table}

During OT reward calculation the best plan $\mu^{\star}$ that matches two trajectories is not time-dependant, since the matching is done regardless of the timestep of each representation, hence, when all the frames are included to the reward calculation, policy can converge to a local minimum where a failed robot trajectory has high matches with expert trajectories that have similar frames throughout different stages of the trajectory. That is why we observe low performance when all the frames are included to the reward calculation, more details are shown in Appendix \ref{sec:reward_details}.

\begin{figure}[t]
    \centering
    \includegraphics[width=\linewidth]{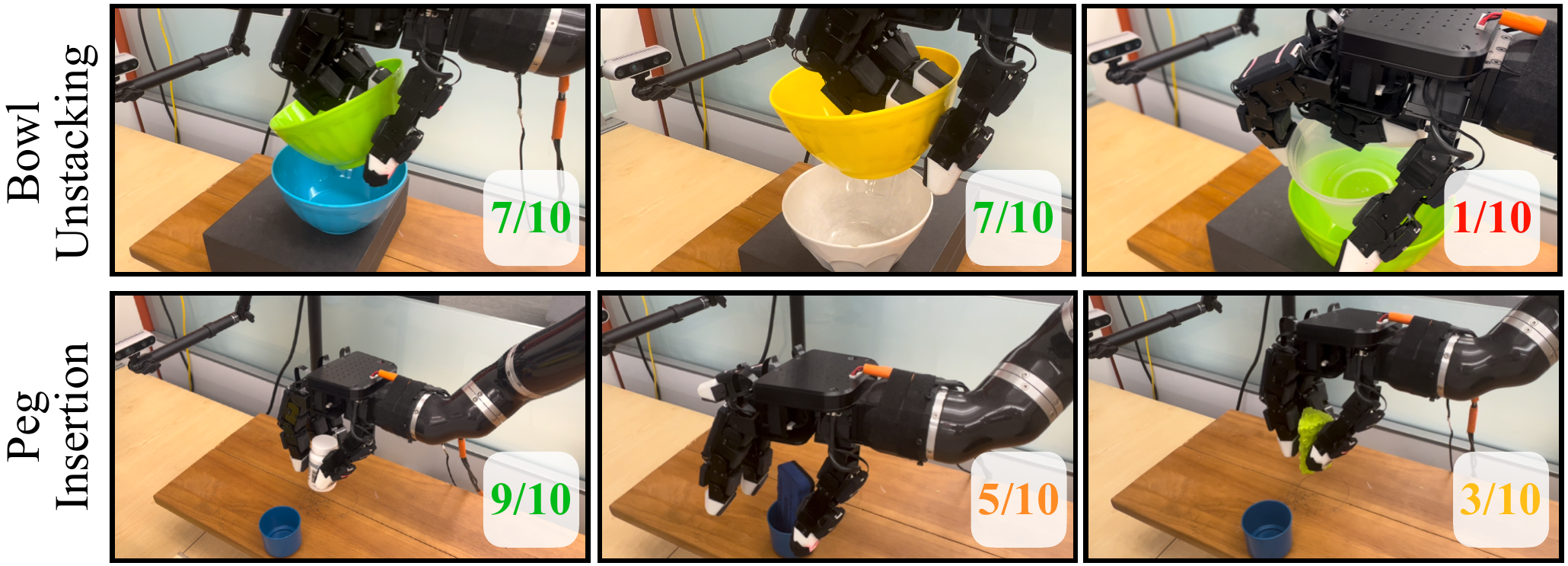}
    \caption{We show success rates of \method{} on a variety of objects not seen during demonstration collection.}
    \label{fig:gen}
\vspace{-10pt}
\end{figure}

\subsection{Does \method{} generalize to new objects?}

We study the ability of \method{} to generalize to unseen objects. For each task, we modify the experiment by replacing one of the objects with a new object with different shape, color, and inertial properties. We run 3 new objects (visualised in Figure \ref{fig:gen}) for the peg insertion and bowl picking tasks, training the policy in the same manner as the original task so it has a chance to adapt. For the bowl unstacking task, we get a success rate of 50\% and for the peg insertion task we succeed 57\% of the time. The policy is able to generalize on some, but not all of the new objects. When the shape or mass of the object changes substantially, the policy is not able to offset the fingertips enough from the base policy to complete the task.

\subsection{Can \method{} be used for long-horizon tasks?}

Due to very large action space of dexterous hands, training long-horizon tasks is a very challenging problem which is why one of the used approaches is to sequence different sub-policies~\cite{chen2023sequential}. In order to sequence sub-policies, each policy learned should also be robust enough for different perturbations coming from each sub-policy.

\begin{figure}[h]
    \centering
    \includegraphics[width=\linewidth]{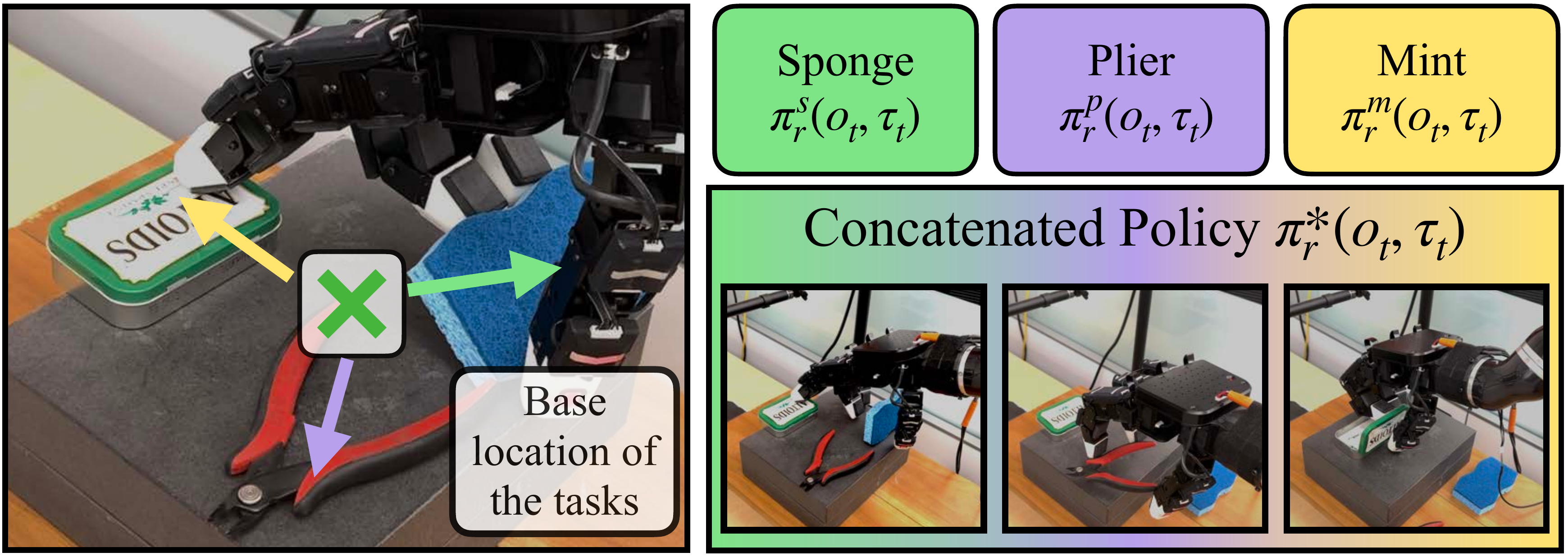}
    \caption{We show an illustration of our long-horizon policy sequencing. \method{} shows robustness when different tasks are sequenced and successfully applies the learned policies separately.}
    \label{fig:multi_task}
\vspace{-10pt}
\end{figure}

We evaluated \method{} to see if it gives robust enough policies to sequence different tasks and give longer horizon tasks. We trained 3 separate policies on our Sponge Flipping, Plier Picking and Mint Opening tasks. We have enabled additional axes on wrist positions during the training and concatenated these three policies during rollout. \method{} manages to separately flip the sponge, pick the plier and open the mint box zero-shot. We illustrate this on Figure \ref{fig:multi_task}.

\subsection{How robust is \method{} to visual perturbations?}
In order to further analyze how image encoders trained in \method{} handles changes in camera view, we ran additional experiments for our task bowl unstacking where we move the camera around 2cm - 15cm with different orientations and trained \method{} with the representations received from those camera positions. We do not collect new expert demonstrations from the new camera positions which causes the episode camera views to gradually drift from the expert.

\begin{table}[h]
\caption{Success rates of \method{} with different camera views.}
\begin{tabular}{ccccc}
\toprule
Positional Variations          & None & ~2cm          & ~2-10cm           & ~12cm+orientation  \\ \midrule
Bowl Unstacking &     9/10      & 6/10          & 2/10          & 1/10           \\ \bottomrule
\end{tabular}
\label{tab:view_results}
\end{table}

We see that with small variations, \method{} is still performative. However, with larger variations, the performance drops significantly as the vision-based representations are not trained to be consistent from multiple-views. This causes calculated rewards to be inconsistent with the success of the robot trajectories which makes the policy harder to train. 

\vspace{-0.02in}
\section{Limitations and Discussion}

In this paper, we introduced \method{}, which leverages tactile feedback for dexterous manipulation through optimal-transport imitation learning. We demonstrated its superior performance compared to visual-only policies, identified challenges related to tactile information in reward calculation, and examined key components. Despite its current strengths, we acknowledge three limitations. First, our observation representation lacks historical context; incorporating a transformer could enhance performance but requires solving the challenge of training with limited demonstration data. Second, performance of \method{} seems very dependant on the camera view due to the matching between the expert and the trajectory. Incorporating tactile to the reward or training more robust visual representations to different camera views could mitigate this. Finally,  the exploration mechanism requires knowing which dimensions in the action space to enable. Automating this process could reduce the need for domain expertise. These areas present exciting opportunities for extending \method{}.

\section*{Acknowledgements}

We thank Aadhithya Iyer, Raunaq Bhirangi, Siddhant Haldar, Jyo Pari and Jeff Cui for valuable feedback and discussions. This work was supported by grants from Honda, Meta, Amazon, and ONR awards N00014-21-1-2758 and N00014-22-1-2773.


\bibliographystyle{IEEEtran}
\small
\bibliography{ref}

\clearpage
\newpage
\appendix
\label{sec:appendix}

\subsection{Model Details}
\label{sec:model_details}

We use DrQv2~\cite{yarats2021mastering} as the reinforcement learning (RL) algorithm to train our policy. The input to the policy is the concatenation of the tactile and image representations. This learner uses DDPG~\cite{lillicrap2015continuous} to maximize the reward function.
We showcase the parameters and details used in Table \ref{tab:hyperparams}

\begin{table}[h]
    \begin{center}
    \setlength{\tabcolsep}{6pt}
    \renewcommand{\arraystretch}{1.5}
    \begin{tabular}{ c c } 
        \hline
        Parameter & Value \\
        \hline
            Optimizer                   & Adam\\
            Learning Rate               & $1e^{-4}$\\
            Standard Dev. Schedule      & $1e^{-1}$\\
            Standard Dev. Clip          & $3e^{-1}$\\
            Critic Target Tau           & $1e^{-2}$\\
            Update Actor Freq.          & 4\\
            Update Critic Freq          & 2\\
            Update Critic Target Freq.  & 4\\
            Batch Size                  & 256 \\
            Replay Buffer Size          & 150000\\
            Exploration Steps           & 1000\\
            Aug. (Image)                & RandomShiftsAug \ $pad=4$ \\
            Expert Frame Matches        & 1\\
            Episode Frame Matches       & 10\\
        \hline
    \end{tabular}
    \end{center}
    \caption{DrQv2 Hyperparameters.}
    \label{tab:hyperparams}
\end{table}

\subsection{Reward Details}
\label{sec:reward_details}

\begin{figure*}
    \centering
    \includegraphics[width=\textwidth]{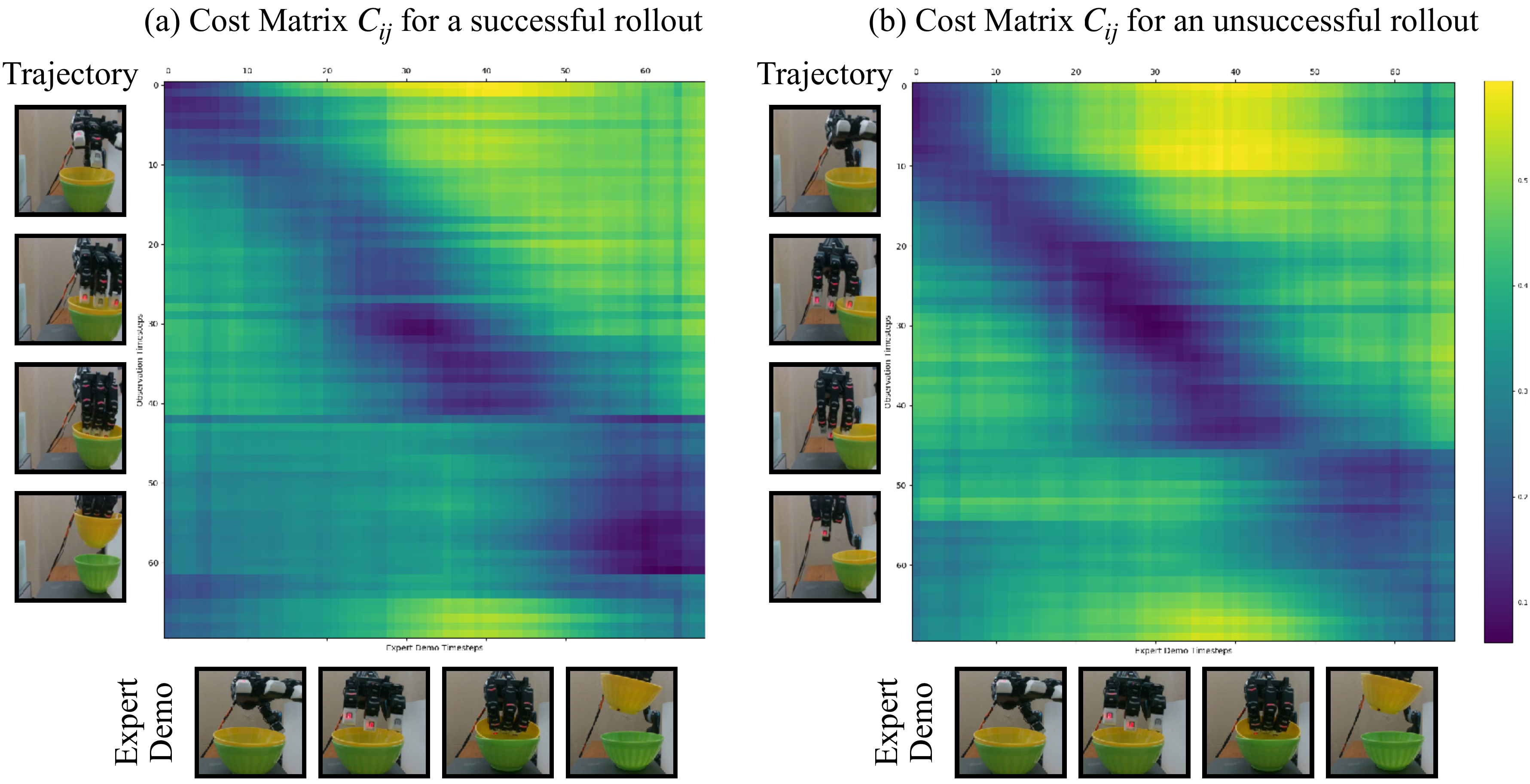}
    \caption{Cost matrix $C_{ij}$ for a failed and a successful trajectory. Darker colors represent low costs and lighter colors represent higher costs. Note the large area of darker colors at the middle of the unsuccessful rollout and the larger area of darker colors at the end of the successful rollout. When OT matching is applied these low cost areas compensate for each other giving an equal reward of -11 for both of these demonstrations. Also note the similarity of the hand pose between the unsucessful and the expert demonstration which explains the similarity of the representations.}
    \label{fig:ap_cost_matrix}
\end{figure*}

In order to further support our decision on choosing the last 10 frames of the episode and the last frame of the expert demonstration, we show the cost matrix $C_{ij}$ when all of the frames are included in Figure \ref{fig:ap_cost_matrix} for a failed and a successful trajectory. Both of these trajectories receive the reward of \textbf{-11} with this way of calculation. 
We observe that when all the frames are included, due to the time independant nature of optimal transport, when there are close representations in different times of the rollouts we receive high scores matches. This problem mostly arises when the hand pose of the trajectory and the expert rollout are similar whereas the objects are in different positions.

In order to tackle this we are only include

\subsection{Training Rollouts}
\label{sec:training_rollouts}

\begin{figure*}
    \centering
    \includegraphics[width=0.8\textwidth]{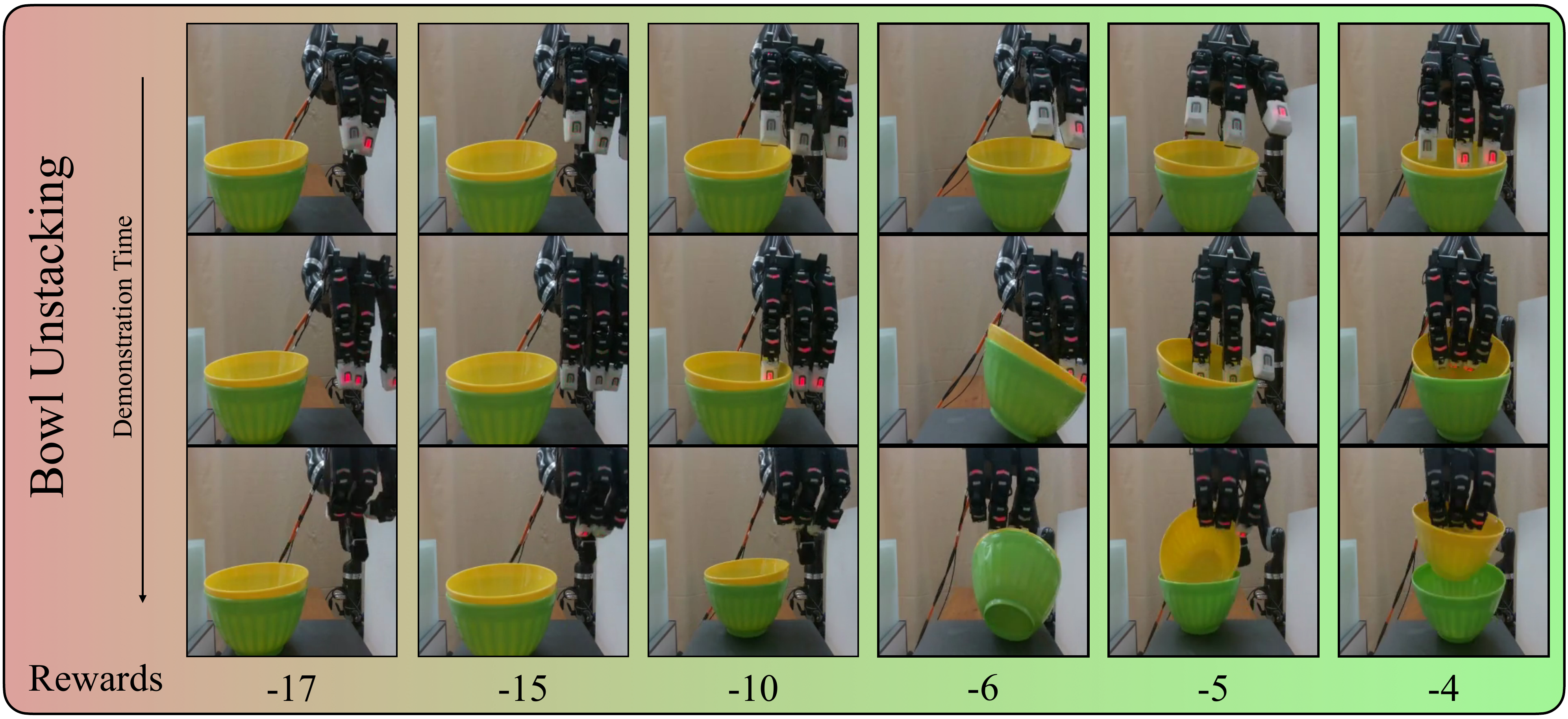}
    \caption{Additional rollouts and corresponding rewards for the Bowl Unstacking task. Note the increase in the reward as the policy improves.}
    \label{fig:ap_bowl}
\end{figure*}

\begin{figure*}
    \centering
    \includegraphics[width=0.8\textwidth]{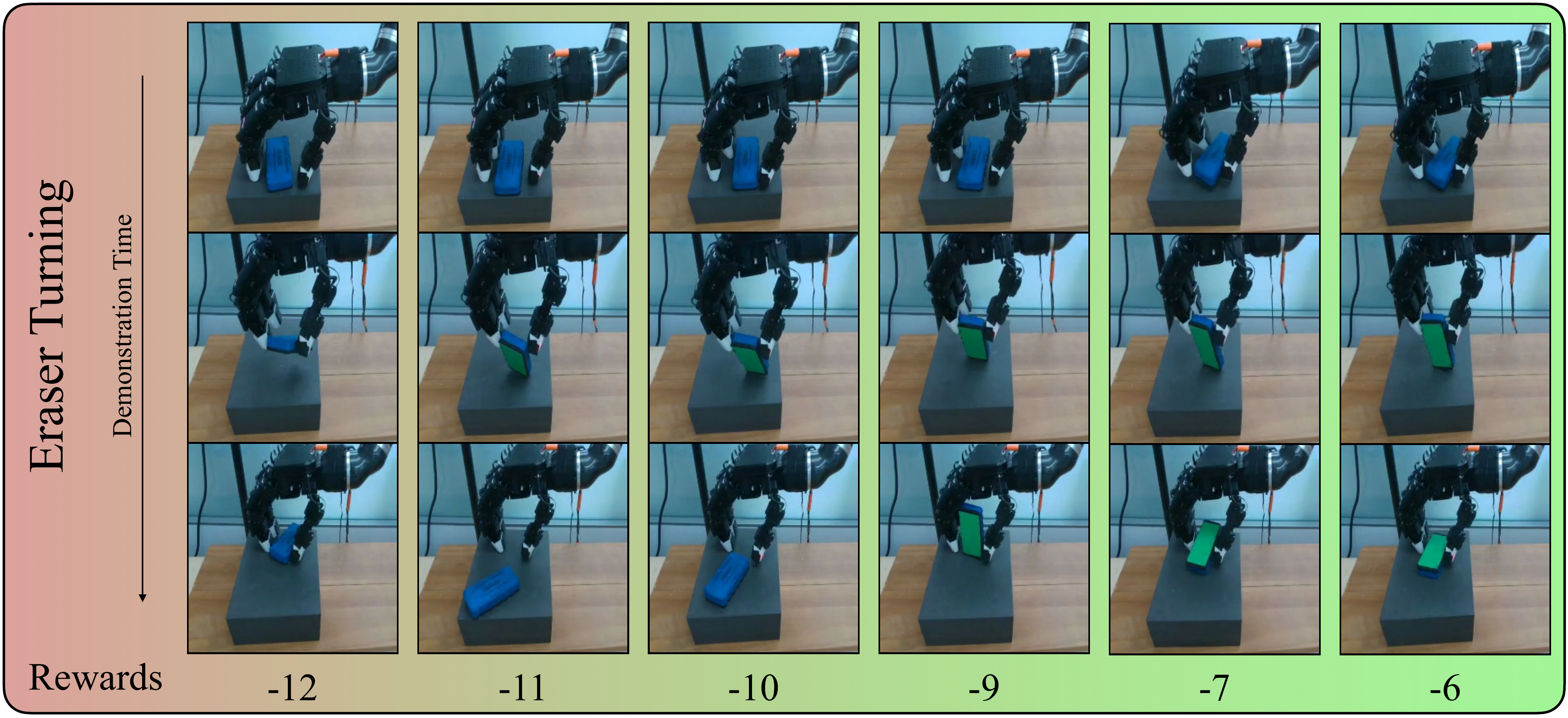}
    \caption{Additional rollouts and corresponding rewards for the Easer Turning task. Note the increase in the reward as the policy improves.}
    \label{fig:ap_eraser}
\end{figure*}

\begin{figure*}
    \centering
    \includegraphics[width=0.8\textwidth]{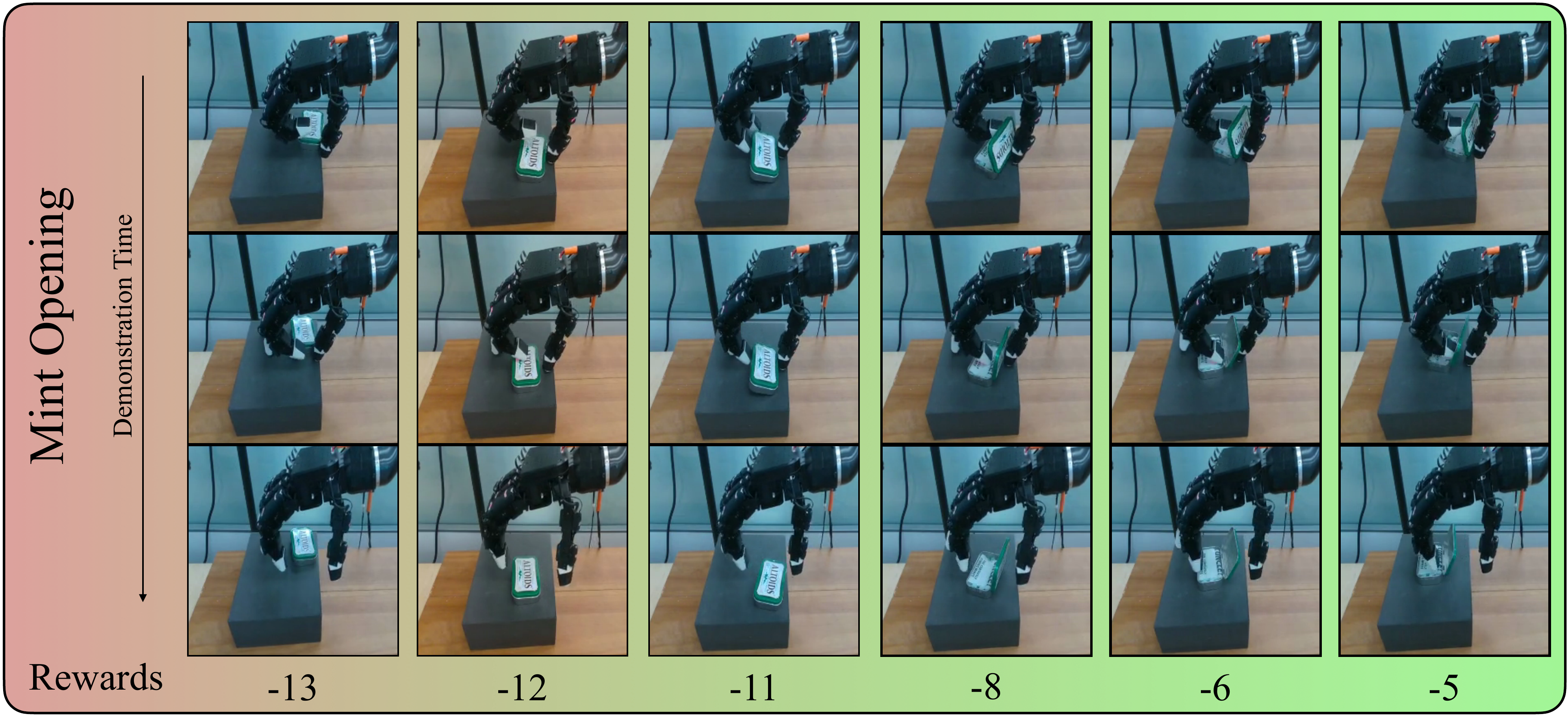}
    \caption{Additional rollouts and corresponding rewards for the Mint Opening task. Note the increase in the reward as the policy improves.}
    \label{fig:ap_mint}
\end{figure*}

\begin{figure*}
    \centering
    \includegraphics[width=0.8\textwidth]{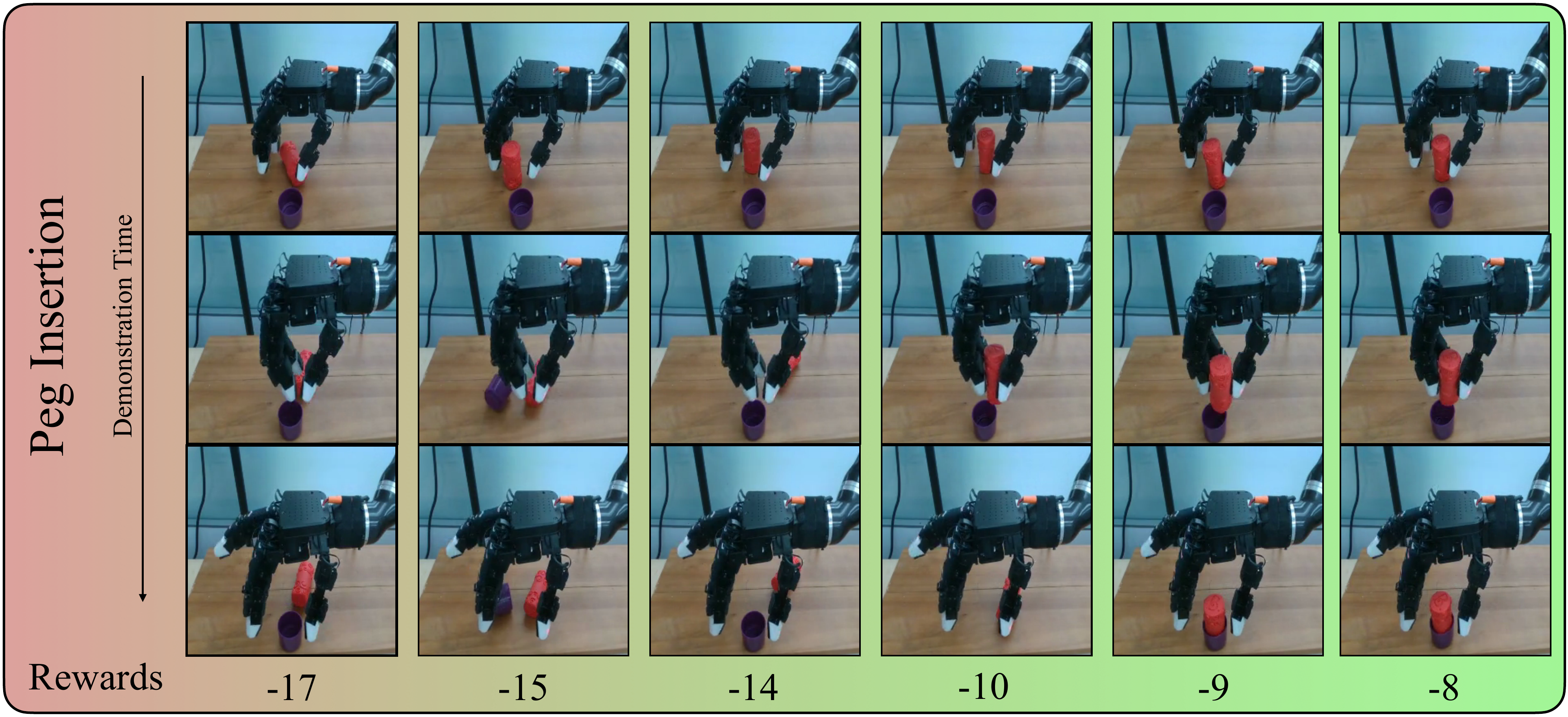}
    \caption{Additional rollouts and corresponding rewards for the Peg Insertion task. Note the increase in the reward as the policy improves.}
    \label{fig:ap_peg}
\end{figure*}

\begin{figure*}
    \centering
    \includegraphics[width=0.8\textwidth]{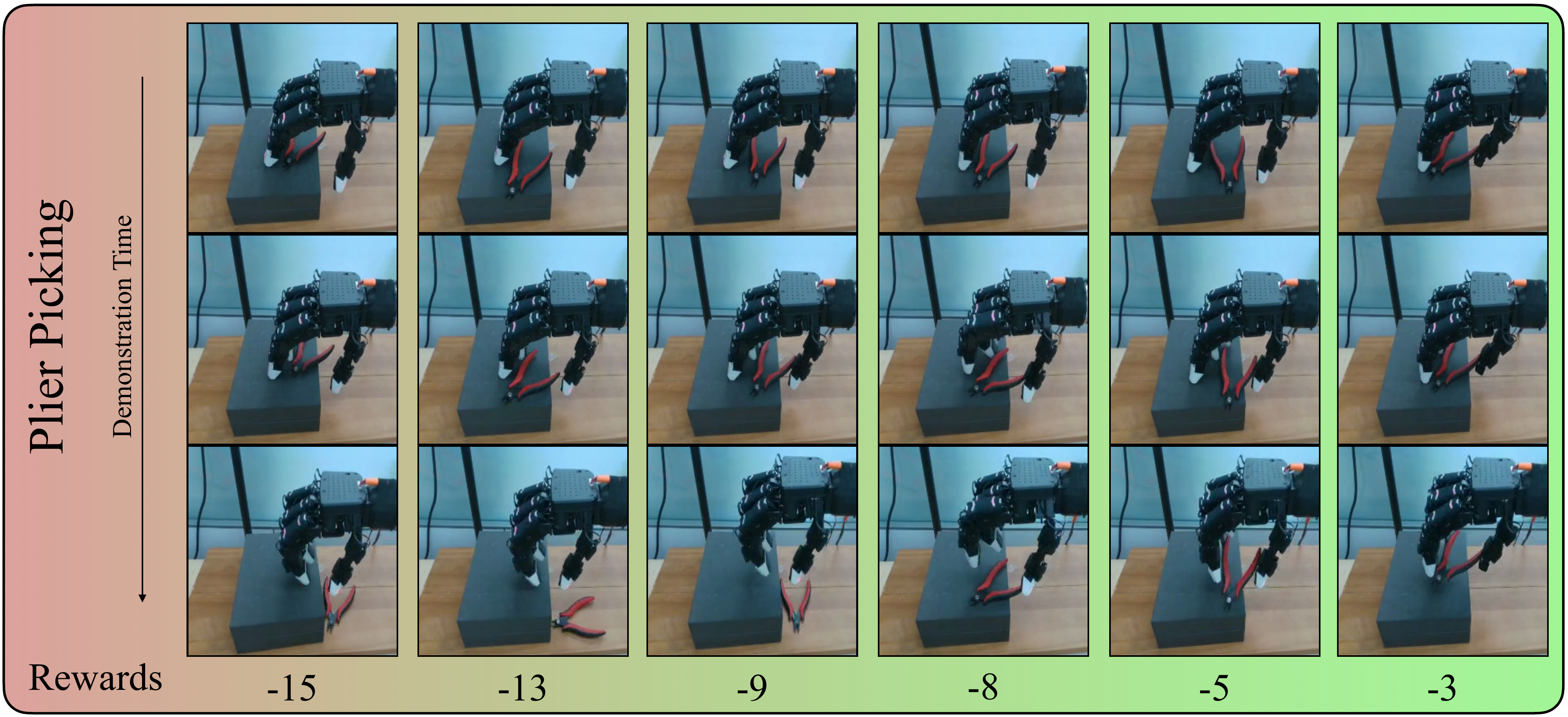}
    \caption{Additional rollouts and corresponding rewards for the Plier Picking task. Note the increase in the reward as the policy improves.}
    \label{fig:ap_plier}
\end{figure*}

\begin{figure*}
    \centering
    \includegraphics[width=0.8\textwidth]{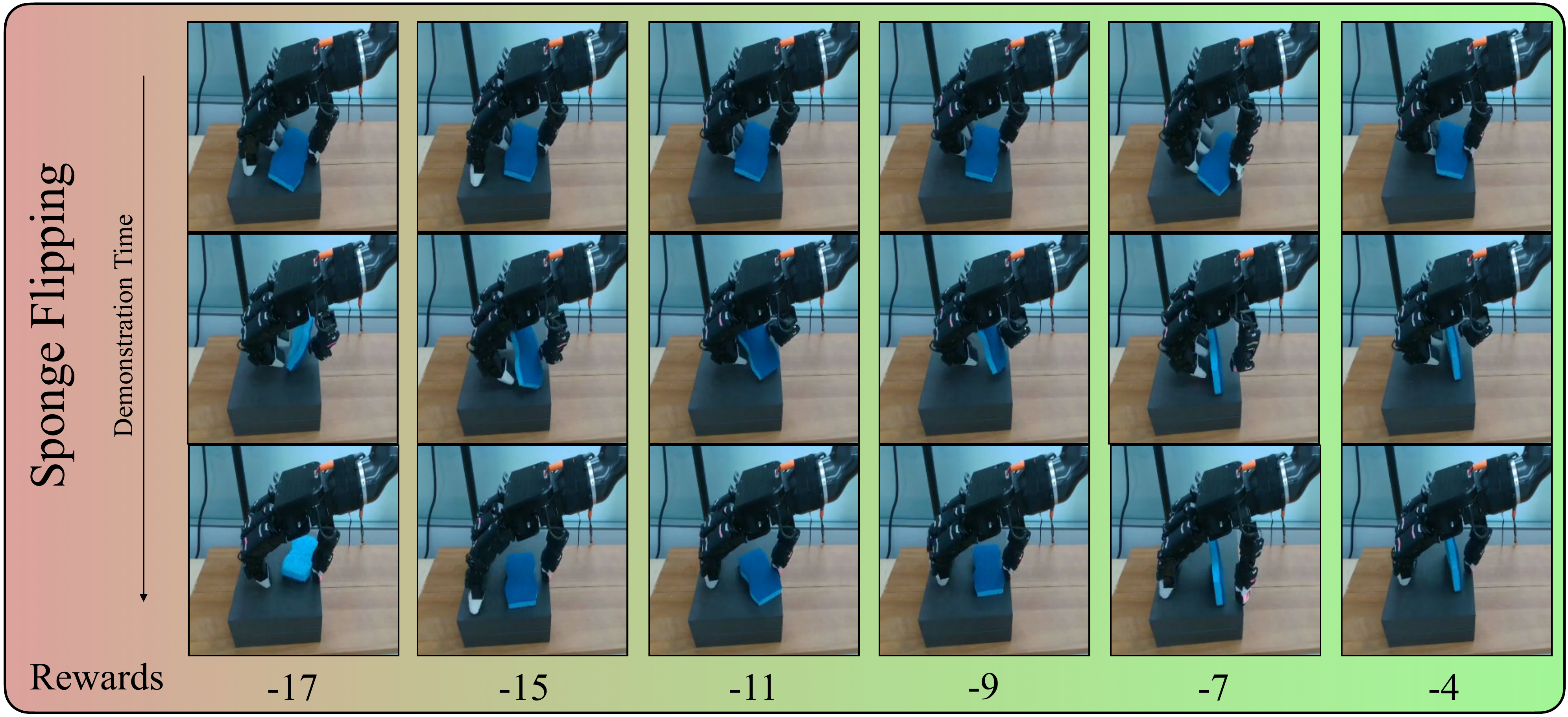}
    \caption{Additional rollouts and corresponding rewards for the Sponge Flipping task. Note the increase in the reward as the policy improves.}
    \label{fig:ap_sponge}
\end{figure*}

We showcase the training rollouts of each task and the corresponding rewards for each rollout in Figures [\ref{fig:ap_bowl}, \ref{fig:ap_eraser}, \ref{fig:ap_mint}, \ref{fig:ap_peg}, \ref{fig:ap_plier}, \ref{fig:ap_sponge}].

\end{document}